\documentclass{article}

\usepackage[dblblindworkshop, final, nonatbib]{neurips_2025}
\workshoptitle{Differentiable Systems and Scientific Machine Learning}

\usepackage[utf8]{inputenc} 
\usepackage[T1]{fontenc}    
\usepackage{hyperref}       
\usepackage{url}            
\usepackage{booktabs}       
\usepackage{amsfonts}       
\usepackage{nicefrac}       
\usepackage{microtype}      
\usepackage{xcolor}         
\usepackage[numbers]{natbib}
\usepackage{amssymb}
\usepackage{amsmath}
\usepackage{xspace}
\usepackage{listings}
\usepackage{minted}
\usepackage{bm}
\usepackage{graphicx}
\usepackage{svg}

\hypersetup{
    colorlinks,
    linkcolor={red!50!black},
    citecolor={blue!50!black},
    urlcolor={blue!80!black}
}

\newcommand{\SOThree}{\ensuremath{\mathrm{SO}(3)}\xspace}
\newcommand{\SEThree}{\ensuremath{\mathrm{SE}(3)}\xspace}

\title{\texttt{scipy.spatial.transform}: Differentiable Framework-Agnostic 3D Transformations in Python}

\author{%
  Martin Schuck \\
  Learning Systems and Robotics Lab\\
  Technical University of Munich\\
  \And
  Alexander von Rohr \\
  Learning Systems and Robotics Lab\\
  Technical University of Munich \\
  \AND
  Angela P. Schoellig \\
  Learning Systems and Robotics Lab\\
  Technical University of Munich \\
}

\begin{document}

\maketitle

\begin{abstract}
  Three-dimensional rigid-body transforms, i.e. rotations and translations, are central to modern differentiable machine learning pipelines in robotics, vision, and simulation. However, numerically robust and mathematically correct implementations, particularly on \SOThree, are error-prone due to issues such as axis conventions, normalizations, composition consistency and subtle errors that only appear in edge cases. SciPy's \texttt{spatial.transform} module is a rigorously tested Python implementation. However, it historically only supported NumPy, limiting adoption in GPU-accelerated and autodiff-based workflows. We present a complete overhaul of SciPy’s \texttt{spatial.transform} functionality that makes it compatible with any array library implementing the Python array API, including JAX, PyTorch, and CuPy. The revised implementation preserves the established SciPy interface while enabling GPU/TPU execution, JIT compilation, vectorized batching, and differentiation via native autodiff of the chosen backend. We demonstrate how this foundation supports differentiable scientific computing through two case studies: (i) scalability of 3D transforms and rotations and (ii) a JAX drone simulation that leverages SciPy’s \texttt{Rotation} for accurate integration of rotational dynamics. Our contributions have been merged into SciPy main and will ship in the next release, providing a framework-agnostic, production-grade basis for 3D spatial math in differentiable systems and ML.
\end{abstract}

\section{Introduction}
Rigid-body rotations and translations underpin modern learning systems in robotics, vision, graphics, state estimation, and control. Their correct use requires precise numerics on Lie groups and algebras, in particular \SOThree and \SEThree, to ensure stability of composition, interpolation, and differentiation. Foundational treatments with emphasis on both the geometric structure and numerically stable parameterizations are well established in mathematics, robotics, and estimation texts \citep{barfoot2017state, Chirikjian2000EngineeringAO, kuipers1999quaternions, murray1994introduction}.

In practice, practitioners either reimplement 3D transform routines to integrate with autodiff and accelerators, which can lead to subtle bugs around singularities, small angles, and representation conventions, or use framework-specific packages. JAX \citep{jax2018github} offers a native SciPy \citep{2020SciPy-NMeth} reimplementation, which is planning to drop support for 3D transformations in a future release \footnote{See \emph{JAX Enhancement Proposal (JEP) 18137} at 
\url{https://docs.jax.dev/en/latest/jep/18137-numpy-scipy-scope.html\#scipy-spatial}}, while PyTorch's \citep{paszke2019torch} ecosystem relies on third-party packages such as \texttt{Kornia} \citep{eriba2019kornia} and \texttt{PyTorch3D} \citep{ravi2020pytorch3d} to provide geometry utilities. The fragmentation into framework-specific packages has so far prevented the development of a unified, portable solution. SciPy's \texttt{spatial.transform} module has provided rigorously tested transformation algorithms in 3D, yet it historically targeted NumPy only and supported at most scalar or 1D stacks of rotations, precluding general broadcasting, autodiff, and accelerator execution.

Orthogonally, SciPy is adopting the Python array API \citep{python2025arrayapi}, a standardized subset of NumPy-like semantics defined by the Python Data API Consortium\footnote{\url{https://data-apis.org/array-api}}. The array API enables library authors to write backend-agnostic numerical code that runs, unmodified, on compliant array implementations such as NumPy, JAX, PyTorch, and CuPy \citep{okuta2017cupy}, thereby inheriting those backends’ accelerators, JIT compilation, and autodiff.

This paper outlines the overhaul of SciPy’s \texttt{spatial.transform} module to conform to the array API and showcases how the new capabilities can be leveraged for differentiable simulation and ML. The revised implementation is framework-agnostic: it executes on CPUs and GPUs/TPUs, is compatible with native autodiff and JIT in the selected backend, and offers numerically robust \SOThree and \SEThree primitives with consistent behavior across libraries. Beyond prior functionality, we add efficient broadcasting across arbitrary leading dimensions, enabling workloads with multiple batch axes common in machine learning. Together, these changes provide a stable, rigorously tested foundation for differentiable 3D spatial computations without reimplementation burden. The overhaul has been merged into SciPy's main branch and will be included in the next release, aligning with the broader effort to reduce parallel reimplementations (e.g., in \texttt{jax.scipy}) as SciPy becomes fully array API compatible.

\section{Framework-Agnostic Rotations and Translations}

\begin{figure}
\centering
    \centering
    \includegraphics[width=\linewidth]{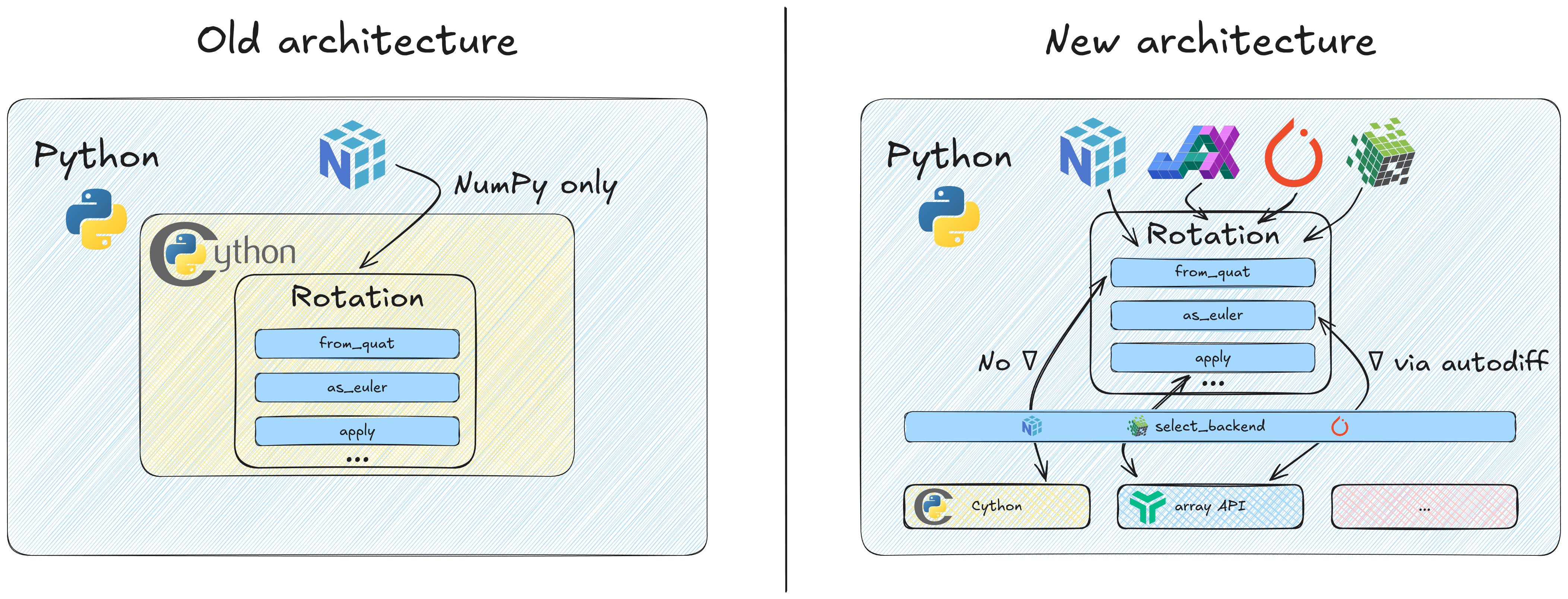}
    \vspace{-1em} 
    \caption{Overview over the changed architecture using \texttt{Rotation} as example. Instead of a pure Cython implementation which is only compatible with NumPy, the classes are now pure Python. Each method delegates computations to a backend that is selected depending on the array type. NumPy arrays are passed to the specialized Cython backend, whereas other array API frameworks use the new, generic backend. This enables advanced capabilities such as differentiation through transforms.}
    \label{fig:architecture}
\end{figure}

\textbf{Previous Design and Limitations} \thickspace \thickspace The original \texttt{spatial.transform} implementation was tightly coupled to NumPy and entirely written in Cython for improved performance. It assumed CPU execution, offered limited broadcasting (only scalars or single-axis stacks), and was incompatible with JIT compilation or automatic differentiation. As a result, downstream users reimplemented core \SOThree/\SEThree functionality per framework, duplicating effort and risking subtle numerical errors.

\textbf{Revised Architecture} \thickspace \thickspace The Cython implementation offers speed advantages over a native Python implementation with NumPy, especially for common use cases such as single rotations or transforms. However, it cannot support arbitrary array API frameworks. To maintain the speedups for NumPy while still supporting other frameworks, we redesigned the architecture of the \texttt{spatial.transform} module. The established object-oriented interface is preserved, but methods now delegate to functional backends selected by the array namespace of their inputs. Fig.~\ref{fig:architecture} provides an overview over the revised architecture. NumPy arrays are processed via the Cython backend, whereas other frameworks use a new reimplementation of rotations and rigid transforms that is fully compatible with the Python array API and thus works for inputs from any frameworks. Each functionality has been rewritten in framework-agnostic, non-branching code, enabling execution on CPUs, GPUs, and TPUs; compatibility with native autodiff and JIT of the chosen framework; and consistent semantics for dtypes, devices, and promotion rules.

\textbf{Data Model and Backends} \thickspace \thickspace Rotations are represented canonically as unit quaternions, rigid transforms as a homogeneous transformation matrix. All constructors and operations support broadcasting across arbitrary leading dimensions, allowing multiple batch axes common in ML workloads without specialized code paths. Third-party arrays that implement the array API are supported out of the box. Advanced users can register custom backends to override operations with optimized versions that are not possible under the limitations of the array API while continuing to use the SciPy interface. These backend are then automatically selected at runtime depending on the array type. This design provides a single, rigorously tested foundation for accurate \SOThree/\SEThree computations across frameworks, eliminating the need for reimplementations.

\section{Case Studies}
In this section, we showcase the capabilities of the new framework-agnostic implementations. First, we evaluate the performance of different frameworks on standard tasks such as rotations and transforms. Then, we demonstrate how the new \texttt{Rotation} class can be used to yield correct analytical gradients for the rotational dynamics using a newly developed differentiable drone simulator written in JAX.

\begin{figure}
\centering
    \centering
    \includegraphics[width=0.95\linewidth]{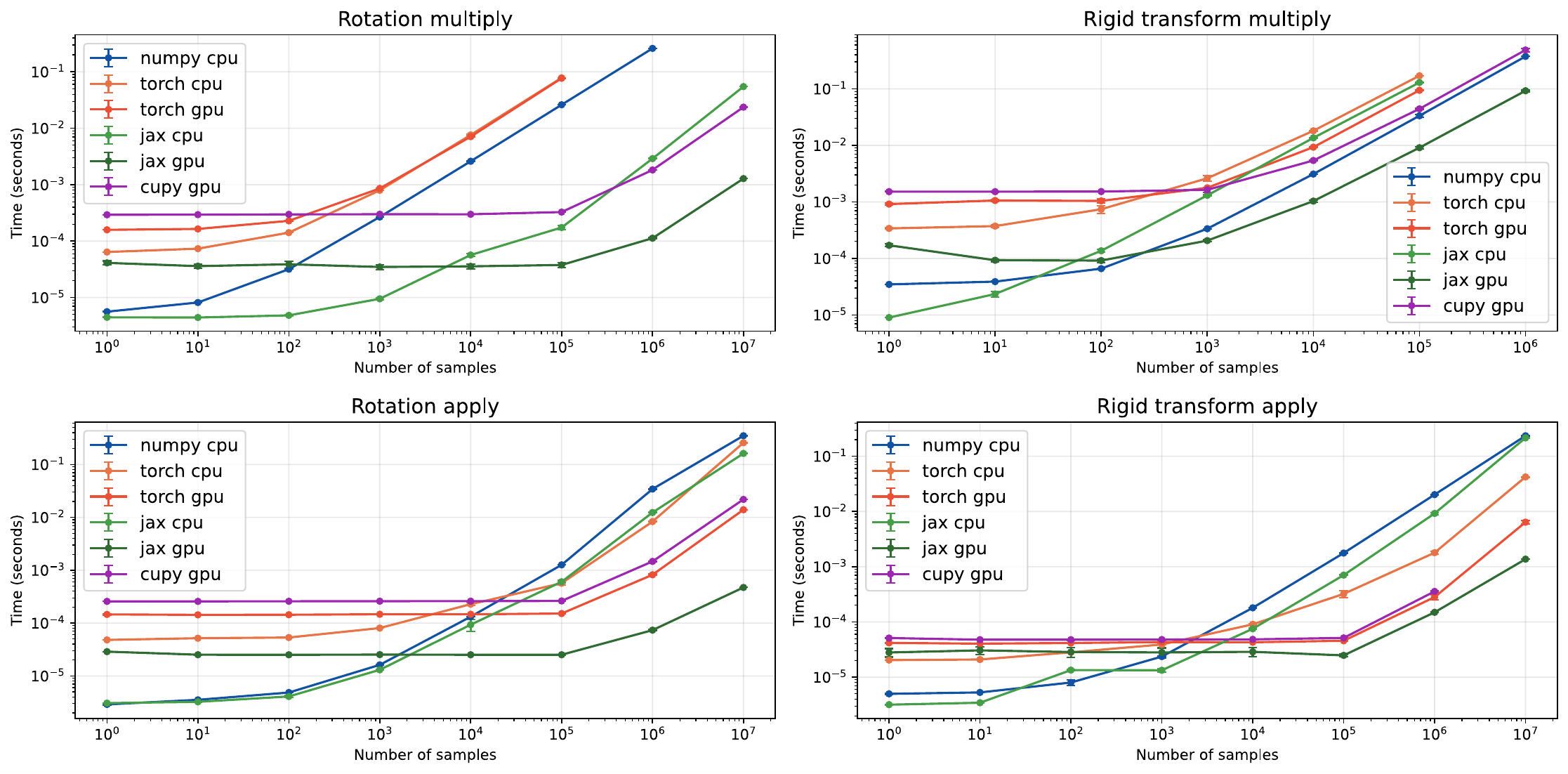}
    \caption{Computation time versus number of samples $N$ for the multiply and apply operations of \texttt{Rotation} and \texttt{RigidTransform} across array API backends. JAX timings are evaluated after JIT compilation. GPU backends incur a fixed overhead at small $N$ but achieve higher asymptotic throughput for large $N$. Notably, JAX often performs on par or better than the custom Cython backend for numpy.}
    \label{fig:benchmark}
\end{figure}

\subsection{Performance per Framework}
We measure throughput for two core operations in the new implementation as a function of the number of rotations or transforms N. The microbenchmarks cover composition and application to 3D points. We evaluate NumPy, PyTorch (CPU and CUDA), JAX (CPU and CUDA), and CuPy. The results can be seen in Fig.~\ref{fig:benchmark}. For small $N$, CPU backends are faster due to lower overhead timings. GPU backends underperform at small batch sizes, then improve as $N$ increases. JAX with JIT compilation attains the highest throughput across the tested workloads; the compiler eliminates array API dispatch overhead and fuses kernels. For large $N$, all GPU backends scale favorably and surpass their CPU counterparts. The crossover depends on the operation and hardware but consistently occurs at sufficiently large batch sizes. NumPy with the Cython-backed SciPy path remains competitive for small $N$ and single transforms, reflecting the optimization targets of the legacy implementation.

\subsection{Differentiable Drone Simulation}
We integrate the new Rotation primitives into a quadrotor simulator to model rigid-body rotational dynamics and differentiate through the simulation. Let $\bm{q}$ be the quaternion describing the current attitude of the drone, $\bm{\omega}_b \in \mathbb{R}^3$ the angular velocity in the body frame, $\bm{J}$ the inertia matrix and $\bm{\tau}_b \in \mathbb{R}^3$ the applied control torques in the body frame. The rotational dynamics are then
\begin{equation}
\bm{\dot{\omega}}_{b} = \bm{J}^{-1} \left( \bm{\tau}_b - \bm{\omega}_b \times (\bm{J} \bm{\omega}_b) \right ), \qquad
\dot{q} = \tfrac{1}{2} (\mathbf{q} \otimes \bm{q}_{\bm{\omega}_b}),
\end{equation}
with the pure-$\bm{\omega}$ quaternion $\bm{q}_{\bm{\omega}_b} = [\omega_x, \omega_y, \omega_z, 0]$. We integrate these dynamics using the group exponential, which preserves unit quaternions by construction:
\vspace{-1em}
\begin{minted}[mathescape,
               linenos,
               numbersep=5pt,
               gobble=2,
               frame=lines,
               framesep=2mm]{python}
    import os
    os.environ["SCIPY_ARRAY_API"] = "1"  # Enable SciPy's array API features
    from jax import Array
    from scipy.spatial.transform import Rotation as R

    def integrate_ang_vel(rot: R, omega: Array, dt: float) -> R:
        return rot * R.from_rotvec(omega * dt)
\end{minted}
Here, the exponential mapping is implemented via \texttt{Rotation.from\_rotvec} and composition uses \texttt{Rotation} multiplication. Because the entire update is expressed with array API operations, autodiff tracks gradients through integration steps in JAX and PyTorch without custom adjoints.

\begin{figure}
\centering
    \centering
    \includegraphics[width=0.32\linewidth,trim={15cm 7.5cm 15cm 0},clip]{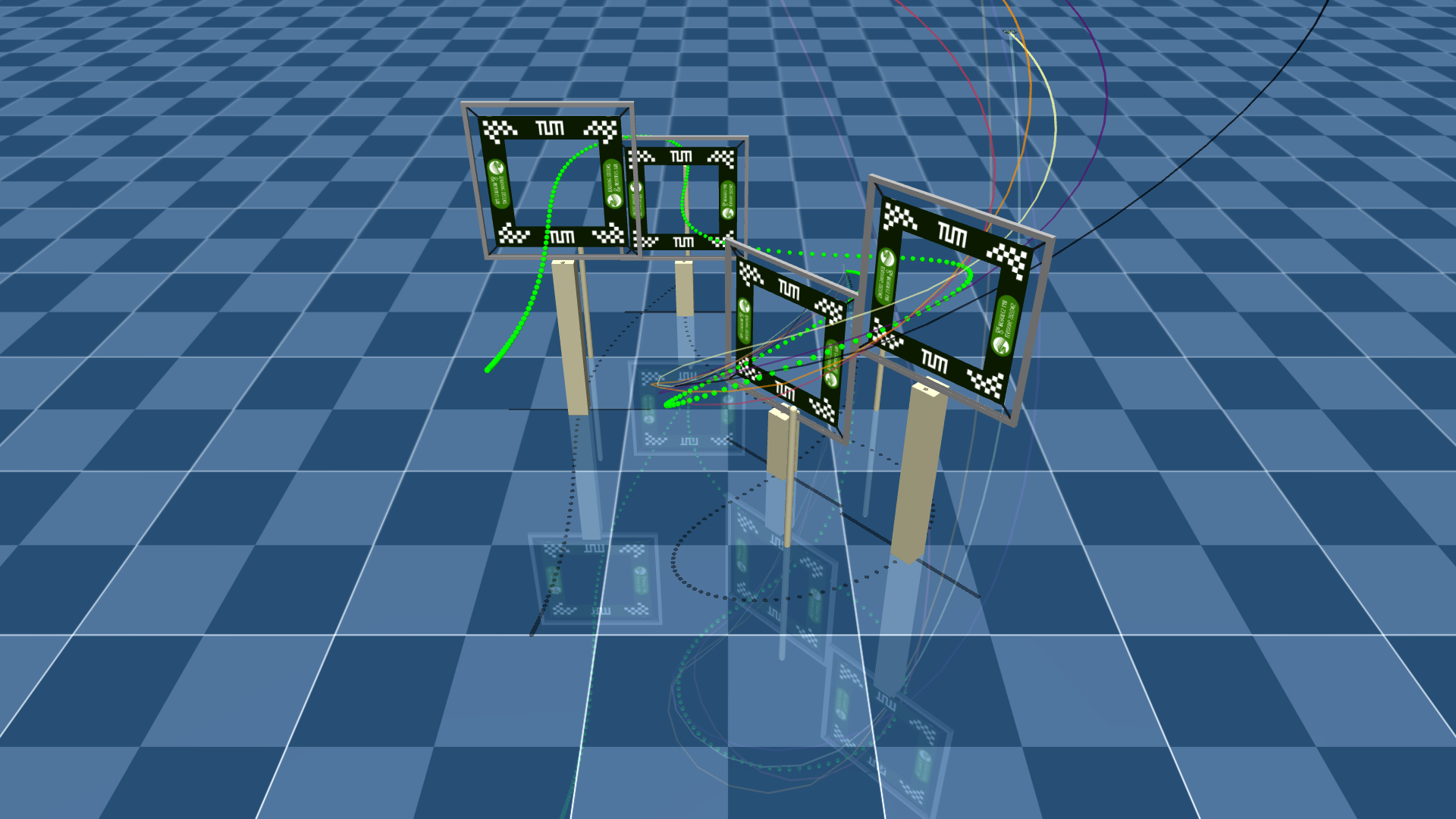}
    \includegraphics[width=0.32\linewidth,trim={15cm 7.5cm 15cm 0},clip]{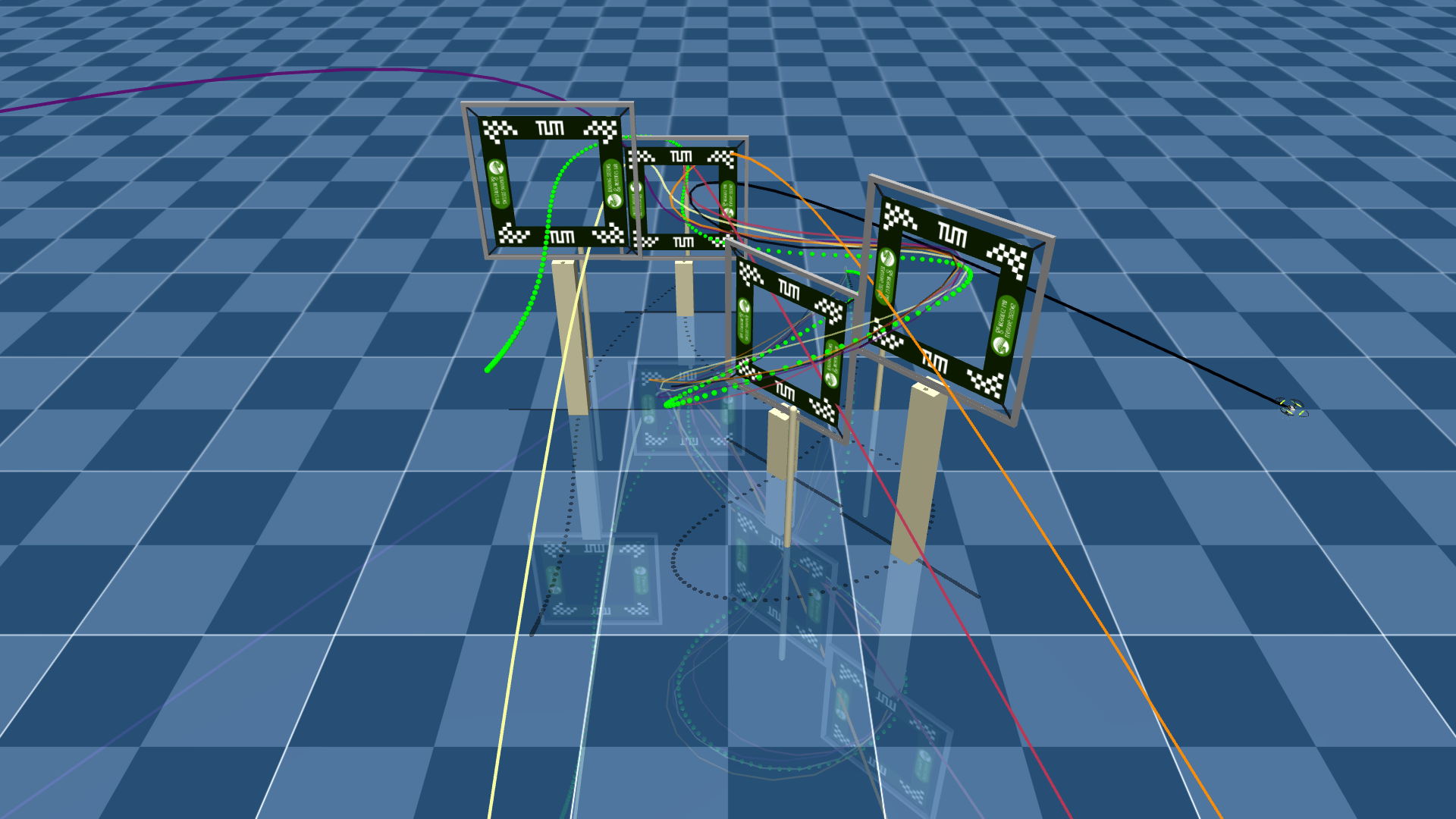}
    \includegraphics[width=0.32\linewidth,trim={15cm 7.5cm 15cm 0},clip]{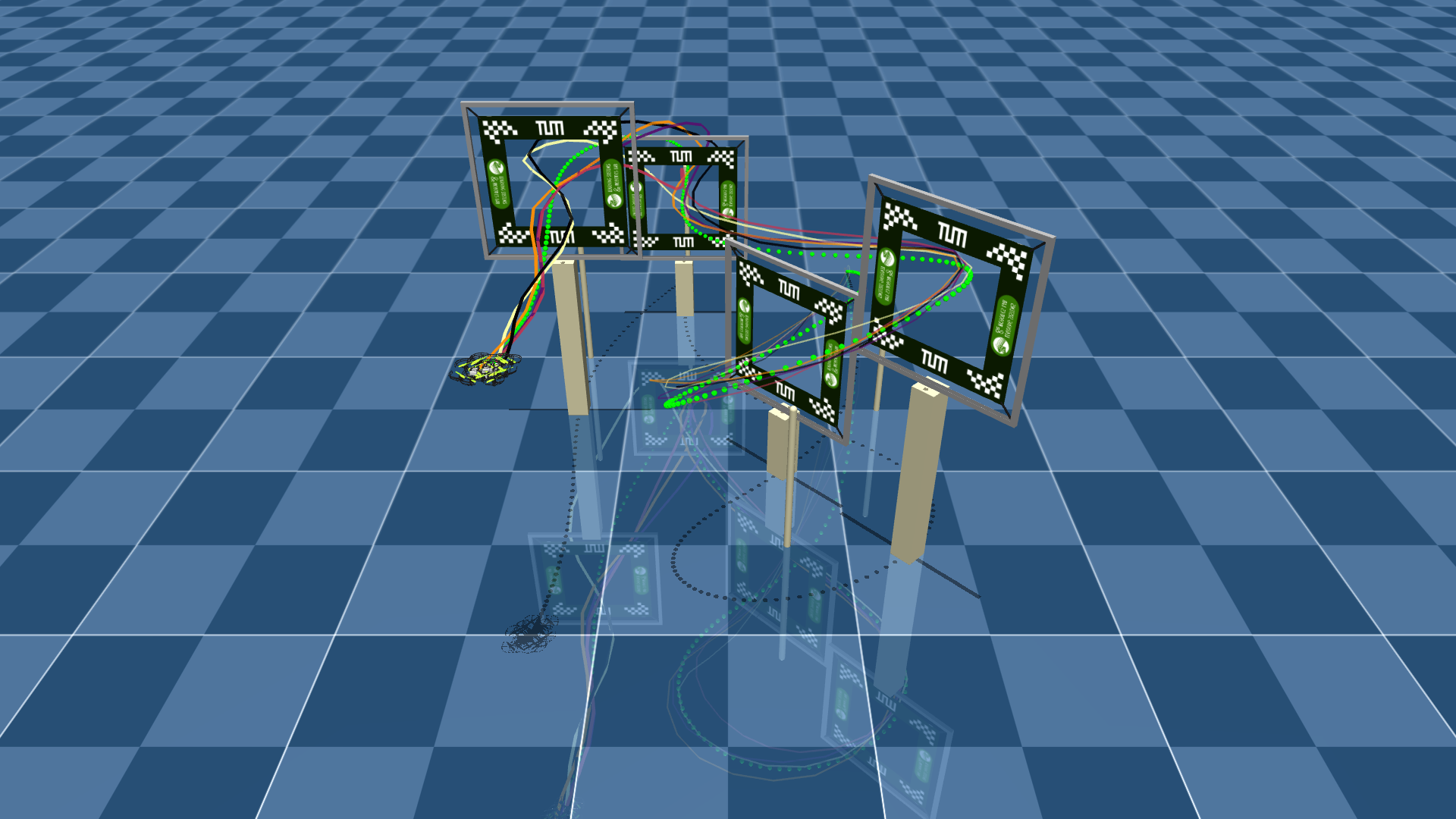}
    \caption{Several drone trajectories optimized with fully differentiable dynamics and drone controllers leveraging \texttt{scipy.spatial.transform.Rotation}. Optimization progress from left to right, reference positions in green. The visualization is based on MuJoCo \citep{todorov2012mujoco}.}
    \label{fig:trajectories}
\end{figure}

We use this differentiable simulator to optimize trajectories for imitation learning in drone flight. Example trajectories can be seen in Fig.~\ref{fig:trajectories}. Gradient-based trajectory optimization adjusts control sequences to minimize a task loss to achieve time optimal flight \citep{romero2022model}, backpropagating through the rotational dynamics integrated with \texttt{Rotation}. This yields optimized trajectories that can, for example, be used as task-aligned supervision signals for policy learning.

\section{Conclusion}
We overhauled SciPy’s \texttt{spatial.transform} to be array API compliant and framework-agnostic while preserving its API, delivering rigorously tested \SOThree/\SEThree primitives that run with NumPy, JAX, PyTorch, CuPy, and others. The new design enables GPU/TPU execution, JIT compatibility, native autodiff, and efficient broadcasting, removing the need for individual, framework-specific reimplementations. Case studies on performance scaling and differentiable drone dynamics demonstrate portability and correctness across devices and frameworks. We invite researchers to incorporate this new capability in their research on differentiable systems whenever they require gradients of rigid transforms or rotations. The implementation is merged and will ship in the next SciPy release. 

\begin{ack}
We thank Lucas Colley, Guido Imperiale and Scott Shambaugh for their input and thorough review of the code. This work was supported by the Robotics Institute Germany under BMFTR grant 16ME0997K, and by the Humboldt Professorship for Robotics and Artificial Intelligence.
\end{ack}

\bibliographystyle{abbrv}
\bibliography{references}

\end{document}